\documentclass{article}

     \PassOptionsToPackage{numbers, compress}{natbib}

     \usepackage[preprint]{nips_2018}

\usepackage[utf8]{inputenc} 
\usepackage[T1]{fontenc}    
\usepackage{hyperref}       
\usepackage{url}            
\usepackage{booktabs}       
\usepackage{amsfonts}       
\usepackage{nicefrac}       
\usepackage{microtype}      
\usepackage[pdftex]{graphicx}

\usepackage[percent]{overpic} 

\title{Topology of Learning in Artificial Neural Networks}

\author{%
  Maxime Gabella \\
  MAGMA Learning \\
  Lausanne, Switzerland \\
  \texttt{maxime@magmalearning.com}
}

\begin{document}

\maketitle

\begin{abstract}
Understanding how neural networks learn remains one of the central challenges in machine learning research. From random at the start of training, the weights of a neural network evolve in such a way as to be able to perform a variety of tasks, like classifying images. Here we study the emergence of structure in the weights by applying methods from topological data analysis. We train simple feedforward neural networks on the MNIST dataset and monitor the evolution of the weights. When initialized to zero, the weights follow trajectories that branch off recurrently, thus  generating trees that describe the growth of the effective capacity of each layer. When initialized to tiny random values, the weights evolve smoothly along two-dimensional surfaces. We show that natural coordinates on these learning surfaces correspond to important factors of variation. 
\end{abstract}


\section{Introduction}

Deep artificial neural networks perform spectacularly on many machine learning tasks, including image classification, speech recognition, translation, and game playing. A neural network can for example be trained to detect the presence of a dog in an image by analyzing a large number of images that have been labeled as ``dog'' or ``no dog,'' each time adjusting its parameters, in particular the weights assigned to edges connecting pairs of neurons. Such a neural network is generally organized in layers that learn increasingly abstract features, from simple combinations of input pixels all the way to full-fledged models of a dog~\cite{DLNature}.

Despite these empirical successes, neural networks are still poorly understood theoretically. One open question is why they generalize so well to unseen samples, whereas the multitude of their parameters would lead one to expect them to overfit the training set~\cite{zhang2016understanding, kawaguchi2017generalization}. Finding the answer would make neural networks more interpretable and provide a principled approach to designing their architecture. 

The main intuition behind this paper is that whatever the structure that arises during training may be, it should be possible to capture it with topology, roughly defined as the mathematical study of qualitative shapes and structures. More specifically, before starting the training the weights are typically initialized randomly and are therefore structureless. However, as the training progresses the weights learn to adjust towards certain distributions of values, whose structure ultimately encodes the knowledge of the neural network about the task at hand. It is the emergence of such learning structures that we wish to exhibit in this paper. 

Our approach is to monitor the evolution of the weights during training. For a fixed layer of the neural network, we consider each of its neurons at each training step as a vector of incoming weights. This provides us in the end with a cloud of points in a high-dimensional vector space. We study the shape of this point cloud by applying techniques from topological data analysis~\cite{carlsson2009topology}. This allows us to represent the evolution of the weights as a graph that encodes the topological structure of the point cloud. The code for this paper uses the Keras library~\cite{chollet2015keras} and is available as a Jupyter Notebook~\cite{CodeMaxime}.

This work was originally motivated by the papers~\cite{DBLP:journals/corr/abs-1810-03234,DBLP:journals/corr/abs-1811-01122}, where topological methods are also used to study the behavior of weights in neural networks. Their approach is however substantially different, as it focuses on exposing circular point clouds representing transformations of $3\times3$ convolutional filters. 

\section{From point clouds to graphs}\label{secGraphs}

In data science, a dataset usually consists of a collection of samples (observations) described by a number of features (variables). The first step in topological data analysis is to represent each sample as a point in a space that has one dimension per feature. The dataset thus takes the form of a cloud of data points in feature space. The main goal is then to study whether this point cloud has some interesting shape, perhaps involving branches or loops. 

\begin{figure}[b]
  \centering
  \begin{overpic}[width=0.9\textwidth]{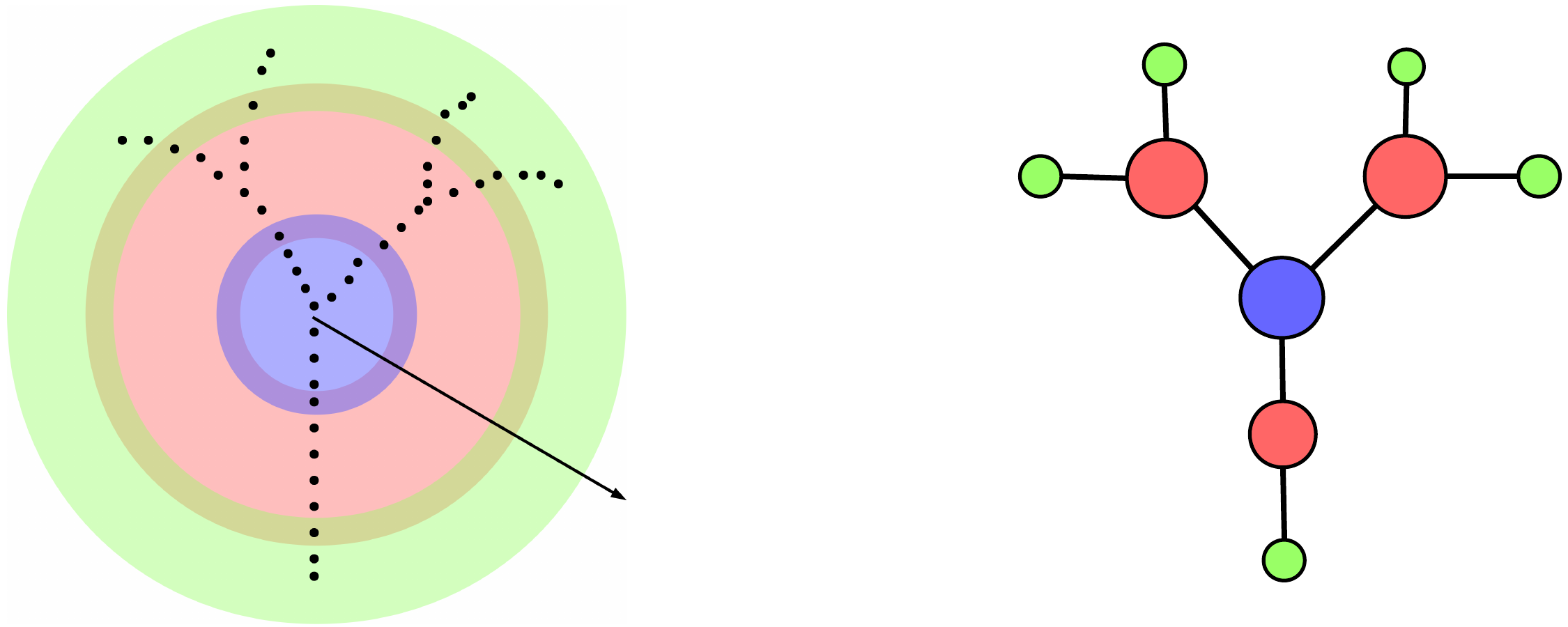}
    \put (0,38) {$X$}
    \put (41,6) {$Z$}
    \put (60,38) {$N(\mathcal{V})$}
  \end{overpic}
  \caption{Mapper algorithm. Here the point cloud has the shape of a tree in $X=\mathbb{R}^2$. The filter function $f: X\to Z$ is given by the distance from the origin, and $Z=\mathbb{R}$ is covered by 3 overlapping intervals (blue, red, green). Clustering each pre-image leads to a covering $\mathcal{V}$ of the point cloud, whose nerve $N(\mathcal{V})$ provides the graph shown on the right. The size of each vertex indicates the number of elements in the corresponding cluster, while edges connect clusters with common elements.}
  \label{MapperAlgo}
\end{figure}

One popular tool to visualize certain topological properties of a dataset is the so-called ``Mapper algorithm''~\cite{Singh2007TopologicalMF} (Figure~\ref{MapperAlgo}). The idea is to construct a graph that captures the topological structure of the original point cloud in the high-dimensional feature space. 

Given a point cloud in a space $X$, the Mapper algorithm first requires to choose a continuous map $f: X\to Z$ to a space $Z$. This map is usually called the \emph{filter function} and can be for example the projection to a subspace of $X$. The next choice is a finite open covering $\mathcal{U} = \{ U_i\}_{i\in I}$ of $Z$. A clustering algorithm is then chosen and applied to each pre-image $f^{-1}(U_i)$ independently, resulting in an open covering $\mathcal{V} = \{ V_j\}_{j\in J}$ of the point cloud. The final Mapper graph (or, more generally, simplicial complex) is then given by the \emph{nerve} of $\mathcal{V}$, that is the set of finite subsets $K$ of the index set $J$ with non-empty intersections: $N(\mathcal{V}) = \{ K \subseteq J | \cap_{k\in K} V_k \neq \emptyset \}$. A subset of $n+1$ indices describes an $n$-simplex. 
In particular, subsets with one element correspond to the vertices of the graph, and subsets with two elements to its edges.
Insightful applications of the Mapper algorithm can be found in~\cite{Lum}.
In this paper, we will use the convenient Python implementation KeplerMapper~\cite{KeplerMapper2019}.

\section{Learning graphs}

\begin{figure}
  \centering
  \begin{overpic}[width=0.99\textwidth]{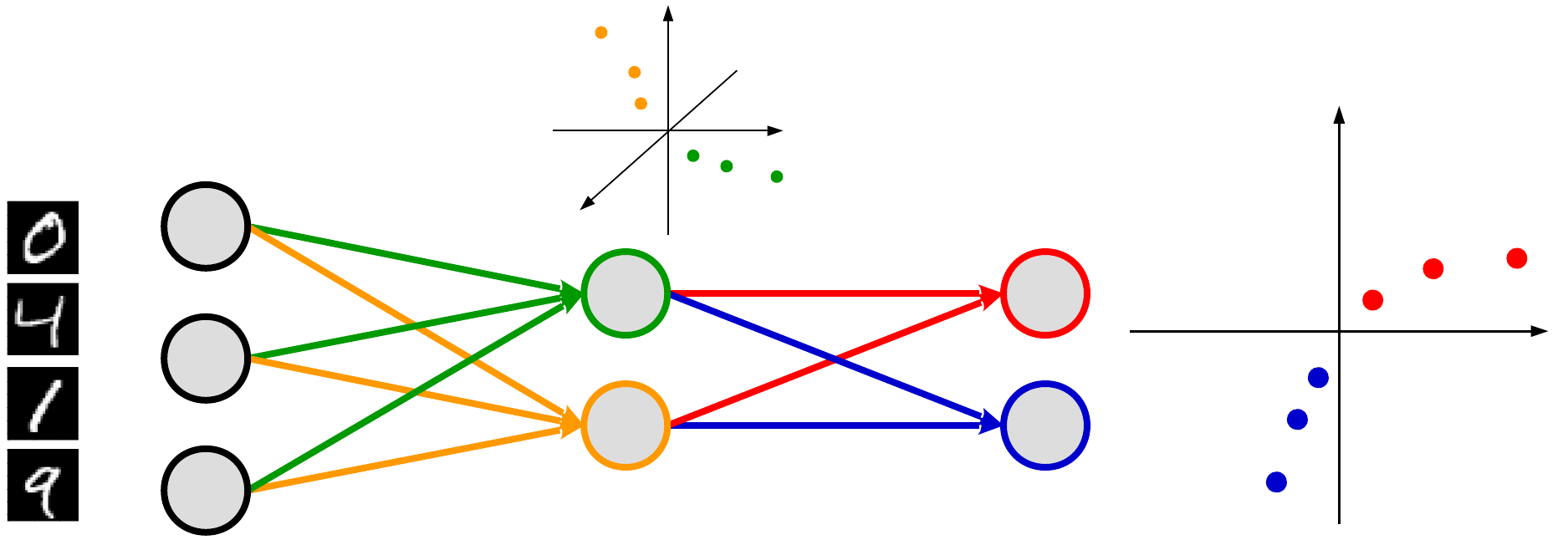}
  \small
    \put(0,23.7){\scriptsize MNIST}
    \put(36.7,2){Hidden}
    \put(63.5,2){Output}
    \put(12.6,19.2){0} \put(12.6,11){1} \put(12.6,2.6){2}
    \put(39.4,15.2){$0'$} \put(39.4,6.7){$1'$}
    \put(43.5,22.6){\tiny$t_1$}\put(46,21.8){\tiny$t_2$}\put(49.2,21){\tiny$t_3$}
    \put(50.5,25.6){\tiny0}\put(43.2,33.3){\tiny1}\put(36.1,20.1){\tiny2}
    \put(87,17.4){$t_1$}\put(91,19.4){$t_2$}\put(96.3,19.9){$t_3$}
    \put(99,12.8){$0'$}\put(84.9,28.2){$1'$}
  \end{overpic}
  \caption{Sketch of a fully connected feedforward neural network. During training, the incoming weights to a given neuron in the hidden or output layer evolve as a point in a space of dimension given by the number of neurons in the previous layer. }
  \label{NeuralNetwork}
\end{figure}
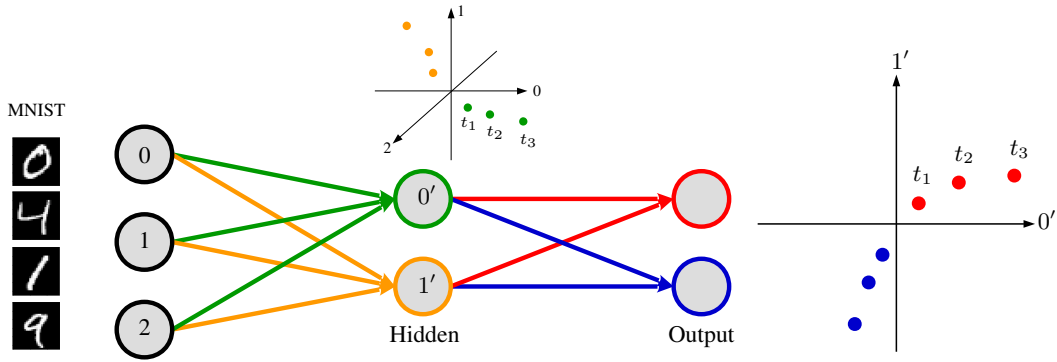

We will now construct Mapper graphs that represent the learning of the weights for each layer of a fully connected feedforward neural network. For a given layer $i$ with $N_i$ neurons, we track the evolution of the incoming weights for each neuron. In other words, we are considering the columns of the weight matrix between layers $i-1$ and $i$. At the beginning of the training the weights are initialized randomly (or to zero), but they evolve to more useful values as they learn. The learning is performed by stochastic gradient descent with minibatches. After each training step (a fixed number of minibatches), we record $N_i$ weight vectors of size $N_{i-1}$, interpreted as $N_i$ points in a space of dimension $N_{i-1}$. By the end of the training we thus have a point cloud of $n_{\mathrm{steps}} \times N_i$ points in $N_{i-1}$ dimensions, whose shape should capture some meaningful aspects of the learning process of the weights (Figure~\ref{NeuralNetwork}). 

Since the number of neuron per layer is typically much larger than two or three, the resulting point cloud lives in a large number of dimensions and cannot be immediately visualized. We will apply standard techniques for dimensionality reduction, like principal component analysis (PCA), to obtain two-dimensional representations. However, as we will see, the results can be fairly involved and hard to decipher. Fortunately, the Mapper algorithm reviewed in Section~\ref{secGraphs} will produce clean graphs that will allow us to easily study topological properties of the evolution of weights during training.

\subsection{Symmetric initialization and branching trees}

We start by studying a neural network with one hidden layer of 100 neurons, trained by stochastic gradient descent on the MNIST dataset of handwritten digits~\cite{726791} (we also trained on Fashion-MNIST and CIFAR-10 with similar results, see the companion notebook~\cite{CodeMaxime}). For simplicity we turn off all biases and only consider weights as parameters. We take sigmoid activation functions for the internal layers, and softmax for the output layer, with the cross-entropy as the loss function. 

Initializing all the weights of a neural network to the same value is generally not recommended in practice~\cite{Goodfellow-et-al-2016}. One of the main guidelines for initialization is indeed to break the symmetry between the weights, so that the neurons can learn different functions. However, since our goal in this paper is not to reach optimal performance but rather to gain some insight into how weights learn, we start by studying the simplest case where all weights are initialized to zero.

\begin{figure}
  \centering
  \begin{overpic}[width=0.3\textwidth]{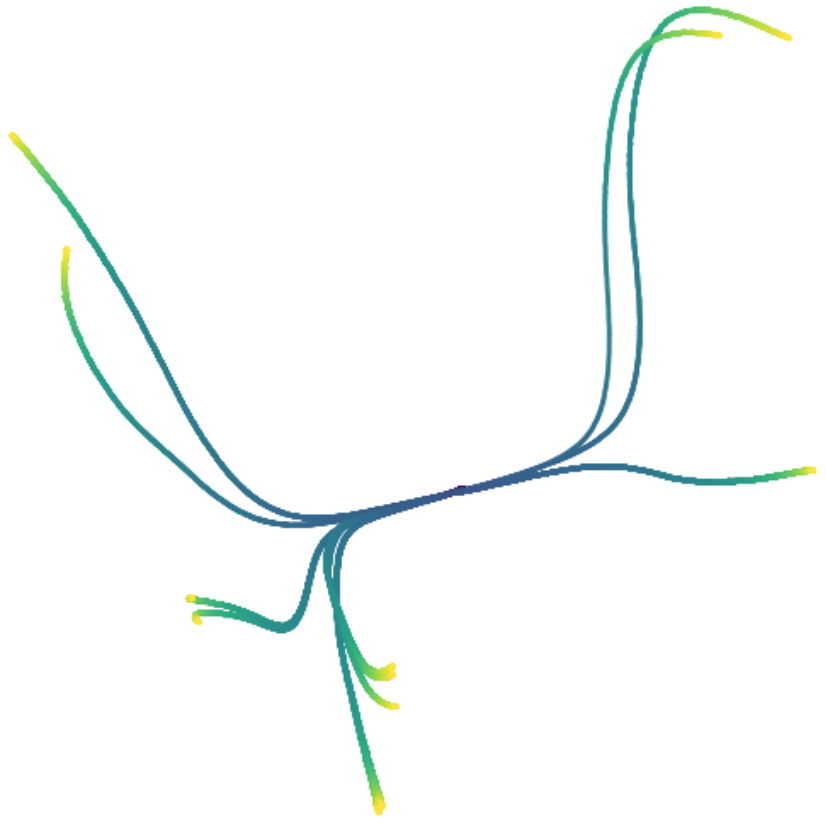}
    \put (0,100) {Hidden layer}
  \end{overpic}
  \hspace{1cm}\hfill
  \begin{overpic}[width=0.3\textwidth]{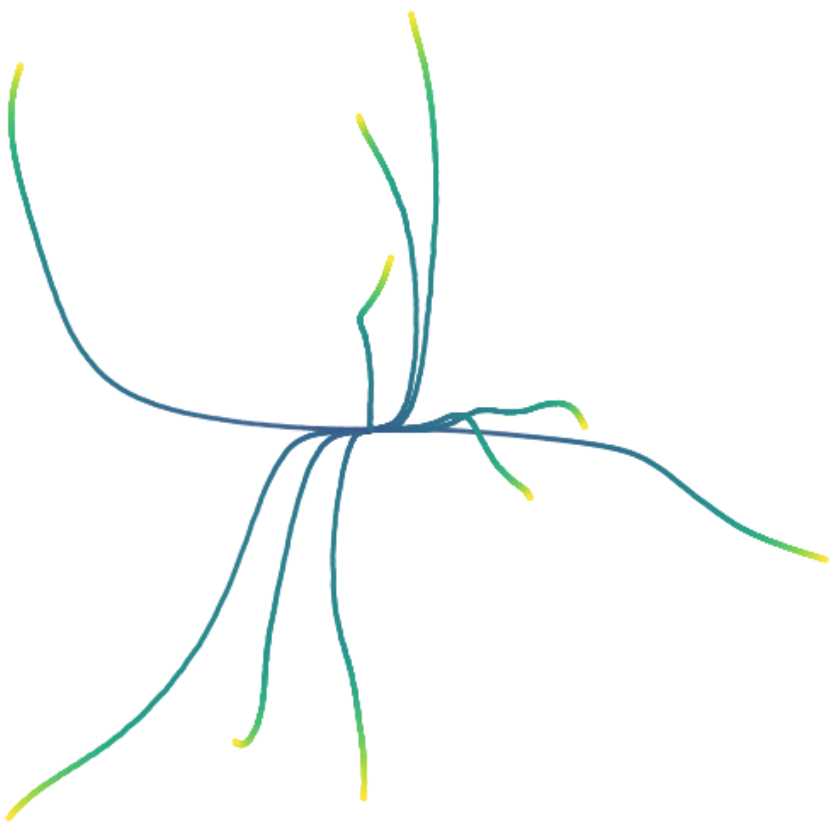}
    \put (0,100) {Output layer}
  \end{overpic}
  \includegraphics[width=0.34\linewidth]{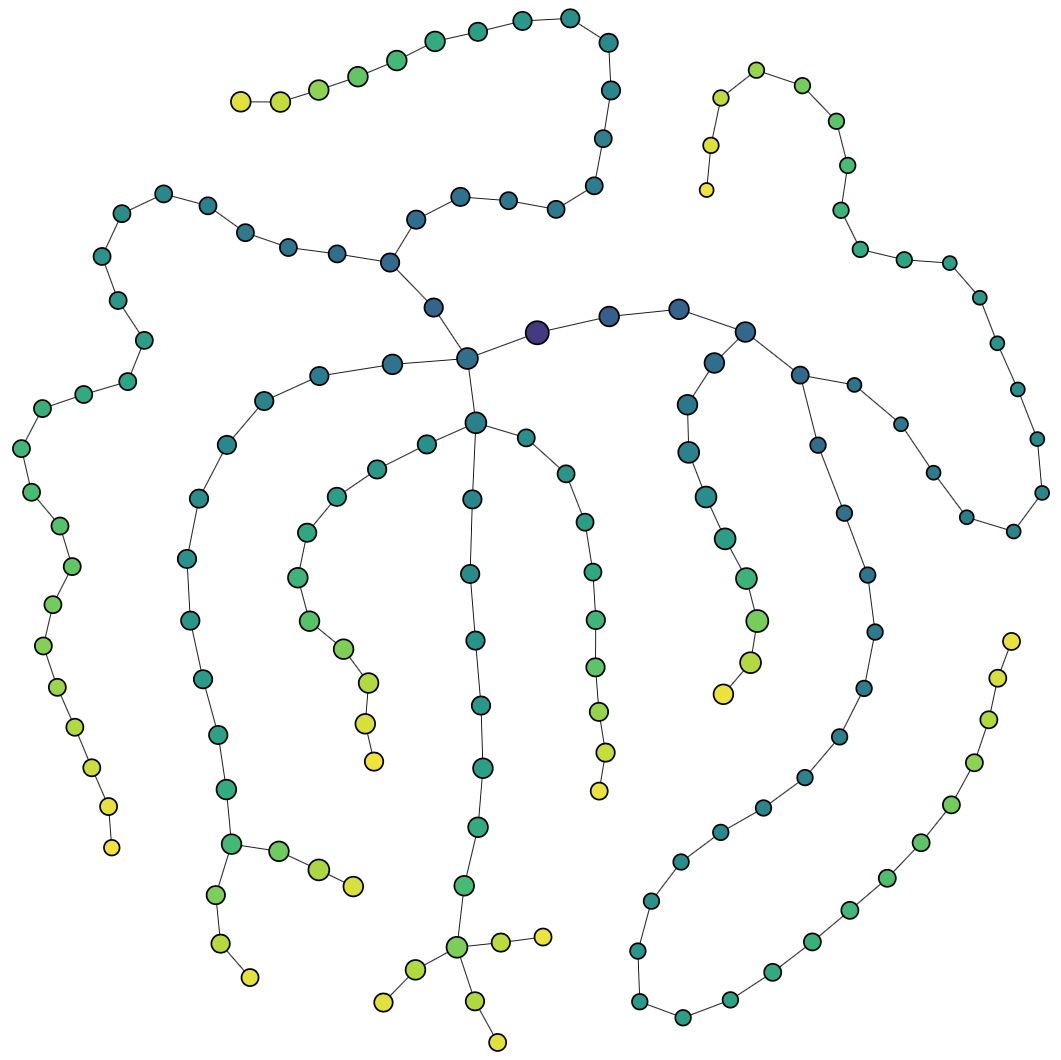}
  \hfill
  \begin{overpic}[width=0.34\textwidth]{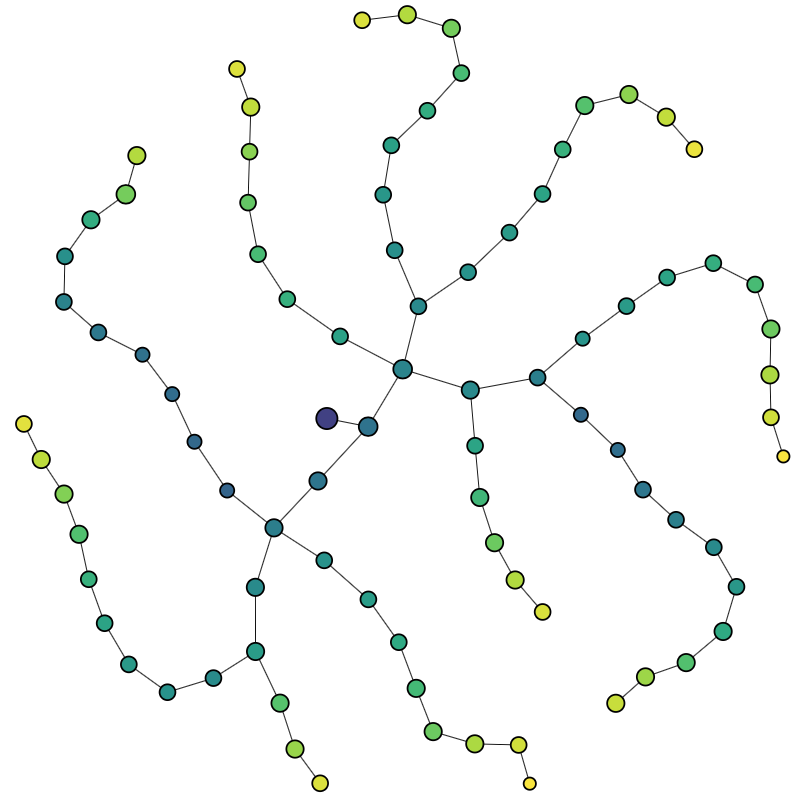}
  \small
    \put (28,50) {Start}
    \put (16,84) {1}
    \put (25,92) {8}
    \put (40,95) {3}
    \put (86,75) {2}
    \put (96,36) {6}
    \put (77,6) {0}
    \put (71,22) {5}
    \put (69,1) {4}
    \put (44,1) {9}
    \put (2,50) {7}
  \end{overpic}
  \caption{Evolution of weights for a neural network with one hidden layer, initialized at zero (colored by training step).
  \emph{Top}: PCA projections. \emph{Bottom}: Learning graphs. 
  \emph{Left}: Hidden layer. \emph{Right}: Output layer (digits are indicated). 
}
  \label{ZeroInit}
\end{figure}
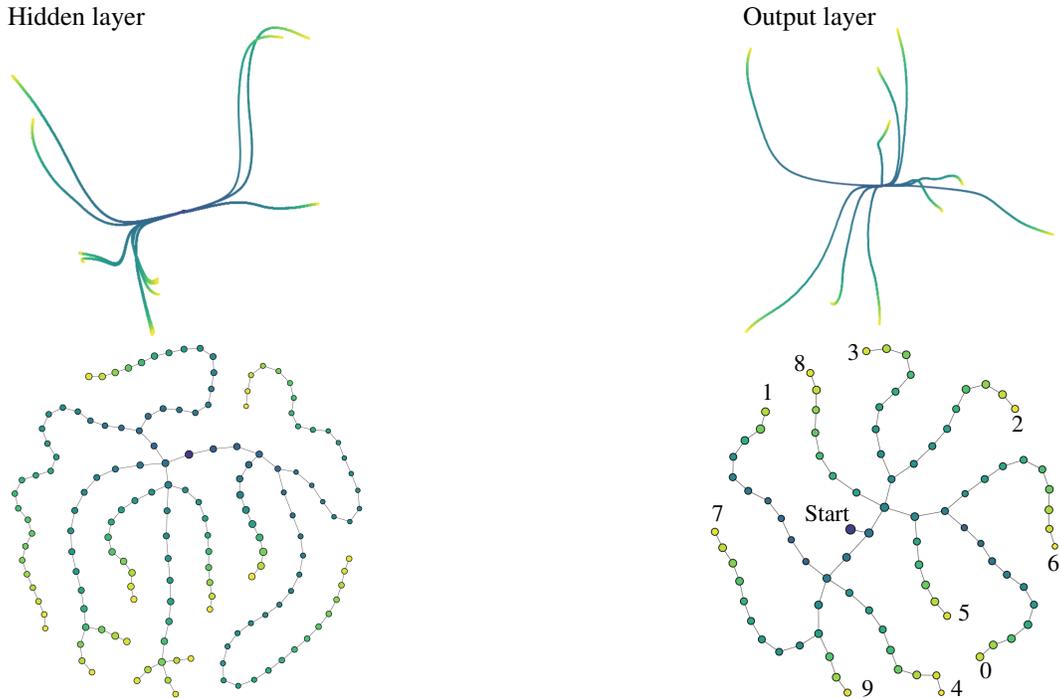

We find that the evolution of the weights during training describes a tree, with subsets of weights occasionally branching off from each other. While this can already be deduced from inspection of the PCA projections, the topological structure of the tree is much easier to grasp from the Mapper graphs (Figure~\ref{ZeroInit}). Here we used the $L^2$~norm as the filter function to take advantage of the fact that the branches grow from the origin, and DBSCAN as the clustering algorithm. 

\begin{figure}[b]
  \centering
  \includegraphics[width=0.95\linewidth]{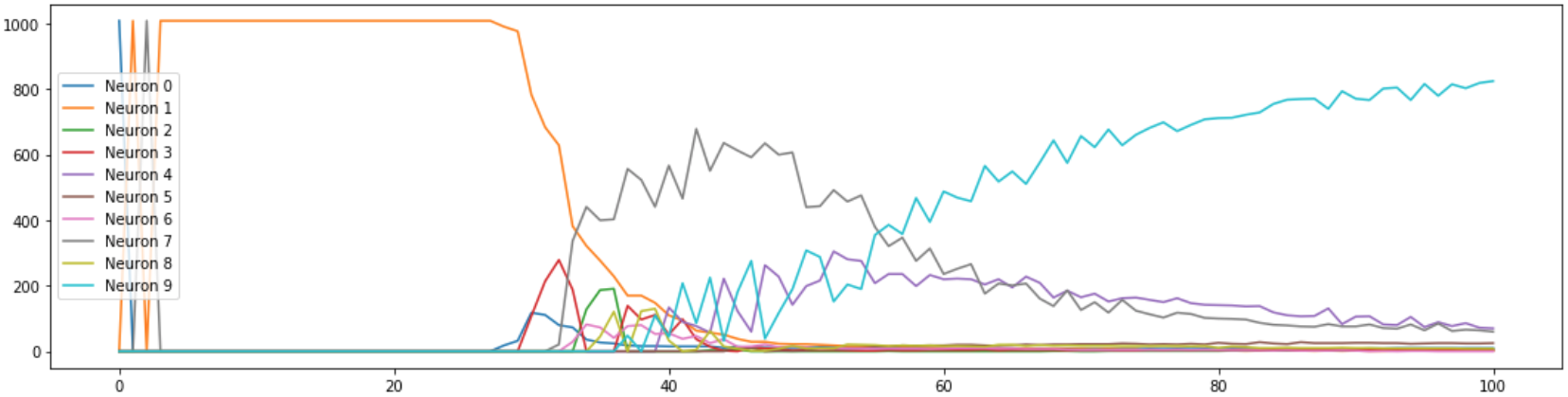}\\
  \hspace{0.41cm}\includegraphics[width=0.835\linewidth]{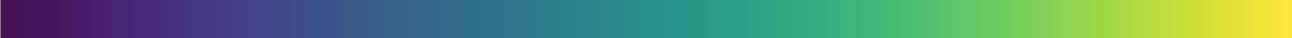}
  \caption{Evolution of the classification of test images of the digit 9. \emph{Horizontal}: percentage of training. \emph{Vertical}: number of classified test samples. The misclassification of nines as sevens loses its prominence around the time when the weights of neurons 7 and 9 branch off in the learning graph.}
  \label{Confusion9}
\end{figure}

Focusing first on the graph for the output layer (bottom right), we see that the weights start by slowly evolving all together along the same trajectory. They then branch off into two trajectories growing in opposite directions. This first branching triggers a steep increase in the model's accuracy. More branchings follow, until there are as many branches as output neurons, one for each of the 10 digits. Interestingly, the discrimination abilities of the model are correlated with the branchings. For example, after a phase where every digit is classified as a 1, the digit 9 gets mostly mistaken for a 7, up to the point when the corresponding branches separate in the learning graph (Figure~\ref{Confusion9}). 

The weights of the hidden layer with 100 neurons evolve in a similar way. After a period of slow evolution (not visible on the graph), they start to branch off recurrently, often at the same time on different branches.  However, in contrast to the case of the output layer, the number of branches at the end of the training does not match the number of hidden neurons. We indeed end up with only 12 distinct branches, which can be taken as an indication of the \emph{effective capacity} of the hidden layer. Additional branchings could of course be produced by extending the training beyond 50 epochs, but it is difficult to find hyper-parameters that lead to the apparition of a maximal number of branches. It would be interesting to determine which conditions lead to maximal branching. 

We note that the shape of the weight trajectories is remarkably constant from one training to the next. This seems to indicate that branching is triggered in a deterministic way. We leave a more complete investigation of the mechanism of branching to future work. 

\subsection{Tiny random initialization and learning surfaces}

\begin{figure}
  \centering
  \begin{overpic}[width=0.3\textwidth]{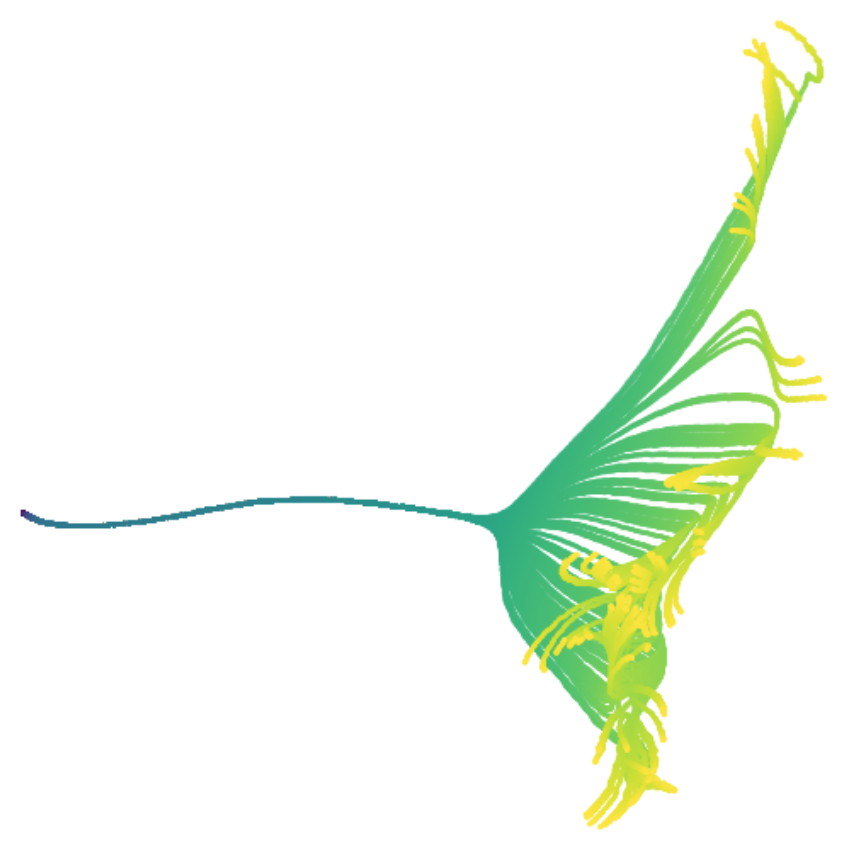}
    \put (0,100) {Hidden layer 1}
  \end{overpic}
  \hfill
  \begin{overpic}[width=0.3\textwidth]{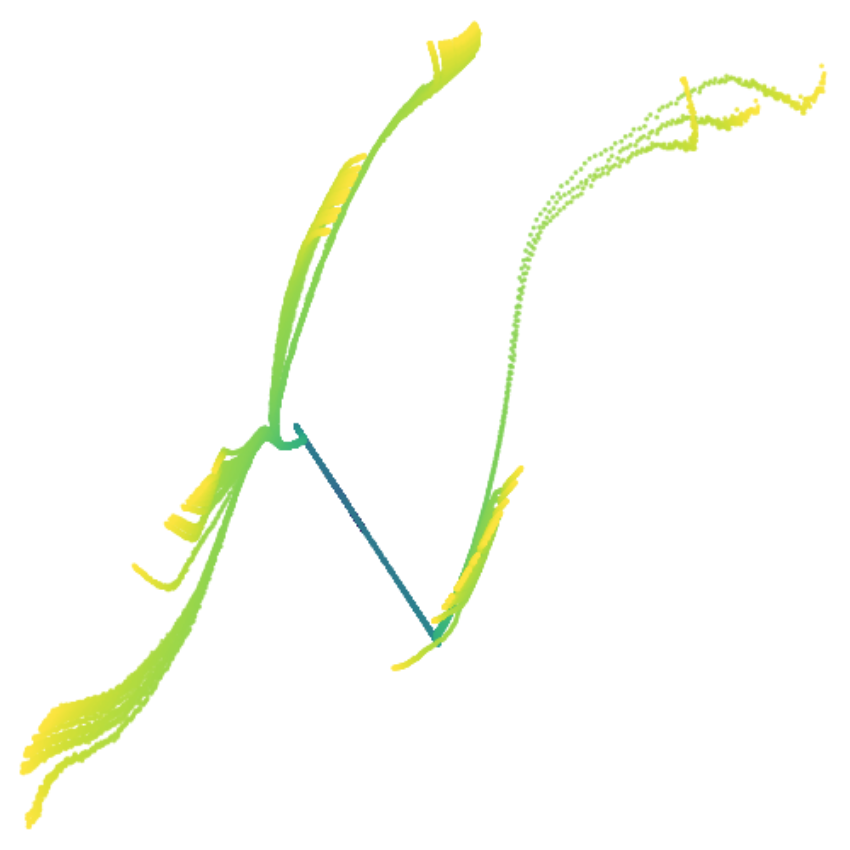}
    \put (0,100) {Hidden layer 2}
  \end{overpic}
  \hfill
  \begin{overpic}[width=0.3\textwidth]{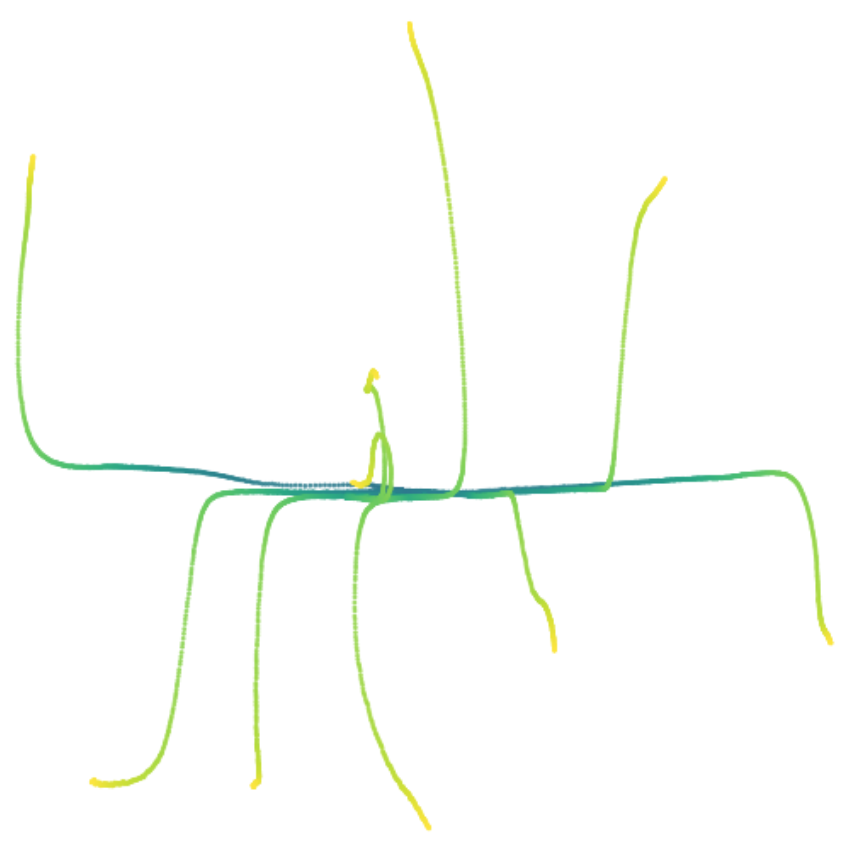}
    \put (0,100) {Output layer}
  \end{overpic}
  \includegraphics[width=0.31\linewidth]{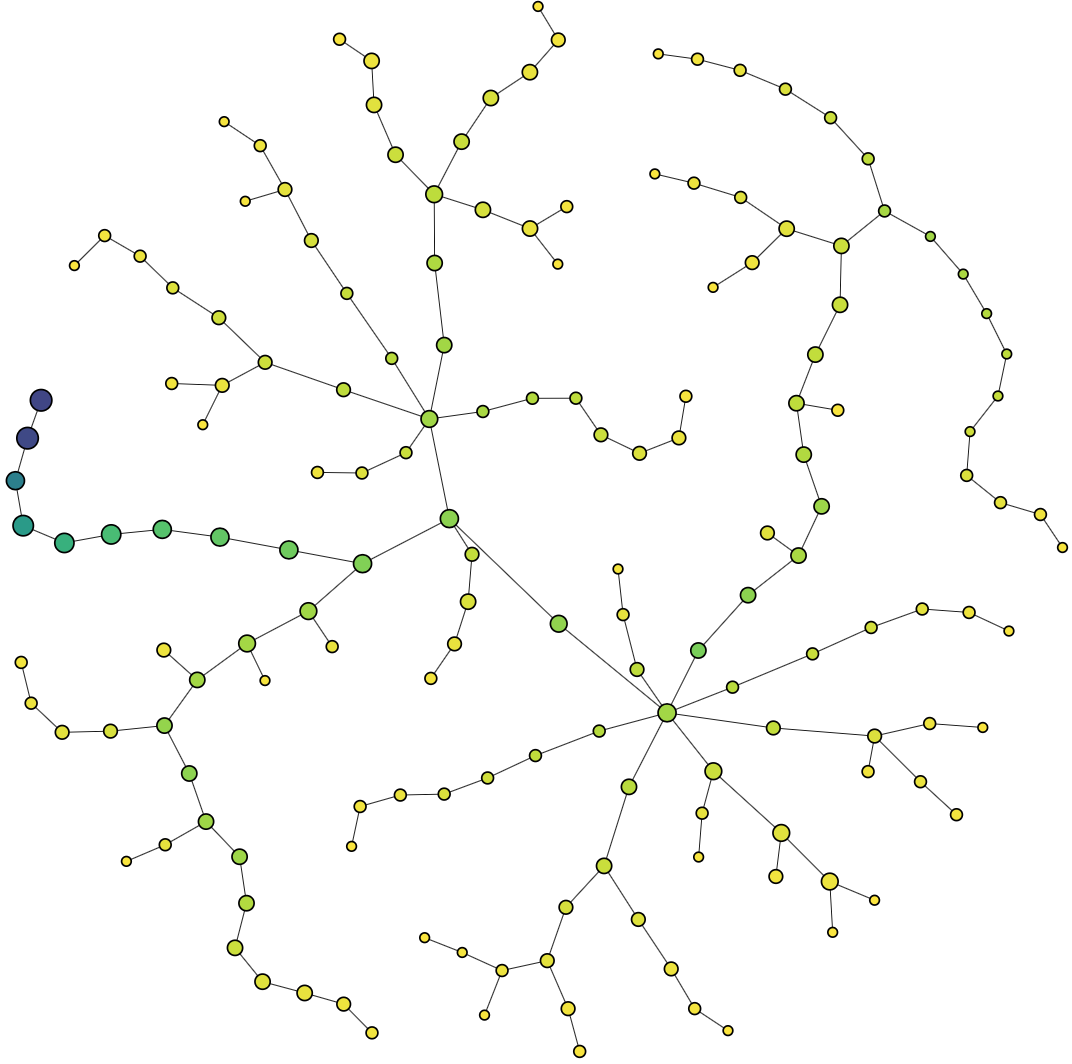}
  \hfill
  \includegraphics[width=0.31\linewidth]{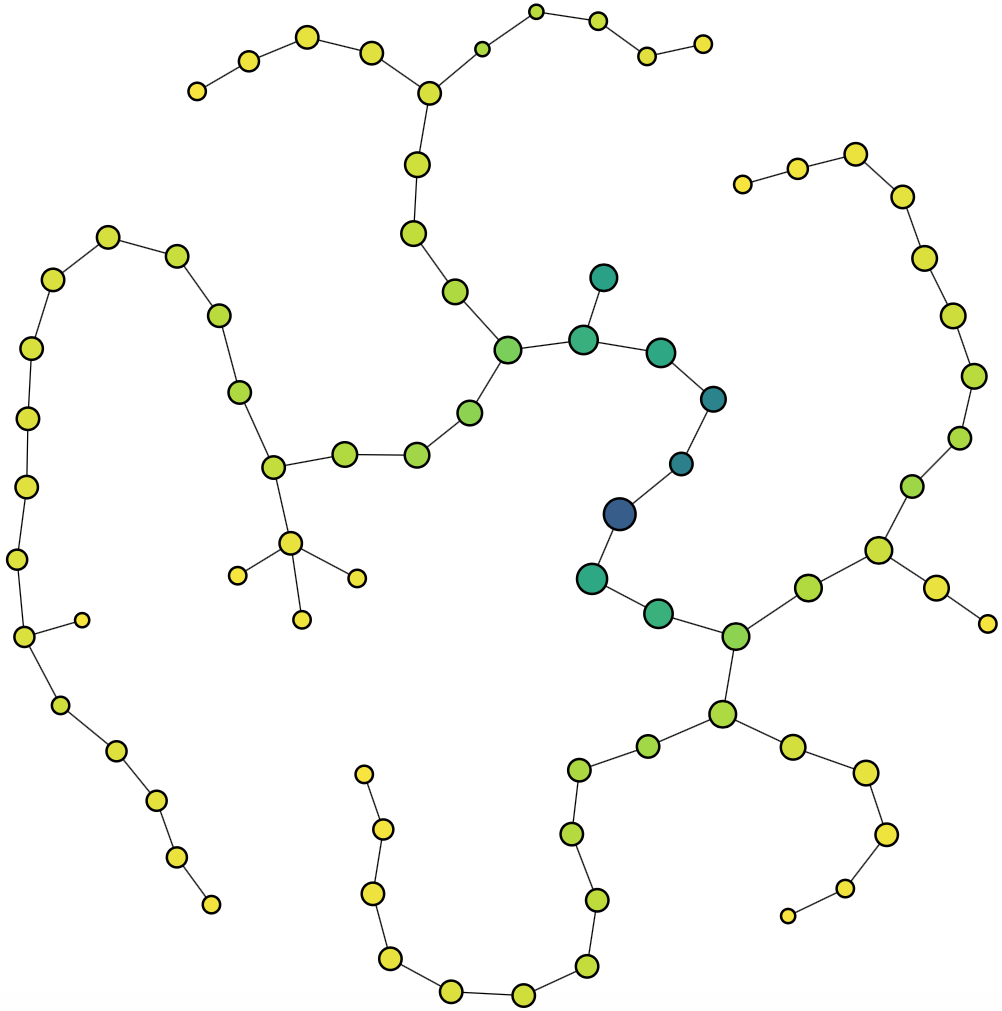}
  \hfill
  \begin{overpic}[width=0.31\textwidth]{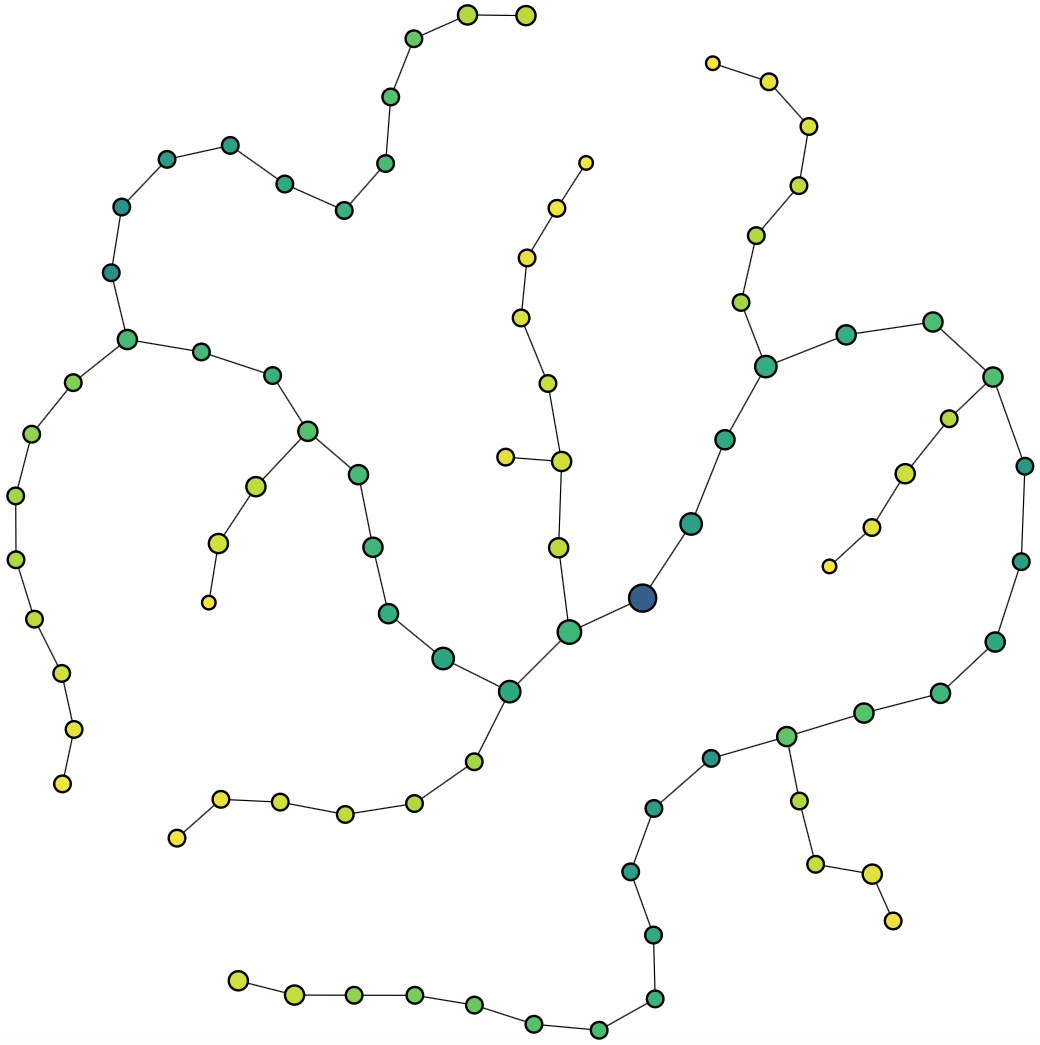}
  \tiny
    \put (53,97) {1}
    \put (44,54) {8}
    \put (50,84) {3}
    \put (66,88) {2}
    \put (88,11) {6}
    \put (18,6) {0}
    \put (75,42) {5}
    \put (15,14) {4}
    \put (3,19) {7}
    \put (18,36) {9}
  \end{overpic}
  \caption{Evolution of the weights of a neural network with two hidden layers, initialized at tiny random values (colored by training step).
  \emph{Top}: PCA projections. \emph{Bottom}: Learning graphs. 
}
  \label{Tiny}
\end{figure}
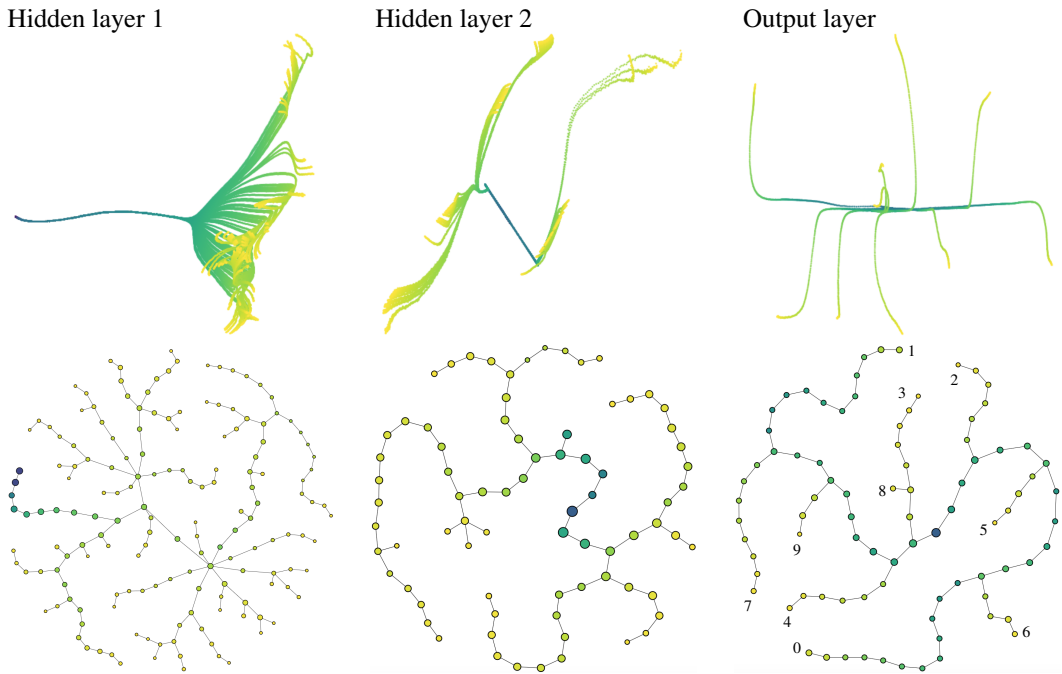

We now turn our attention to the evolution of weights when they are initialized at tiny random values. We consider a neural network with two hidden layers of 100 neurons each and sigmoid activation functions. The weights are initialized with a normal distribution with mean $\mu =0$ and standard deviation $\sigma = 10^{-6}$. We train the network on MNIST (see~\cite{CodeMaxime} for Fashion-MNIST and CIFAR-10) and record the evolution of the weights throughout training for each layer (Figure~\ref{Tiny}).

\begin{figure}
  \centering
  \includegraphics[width=0.37\linewidth]{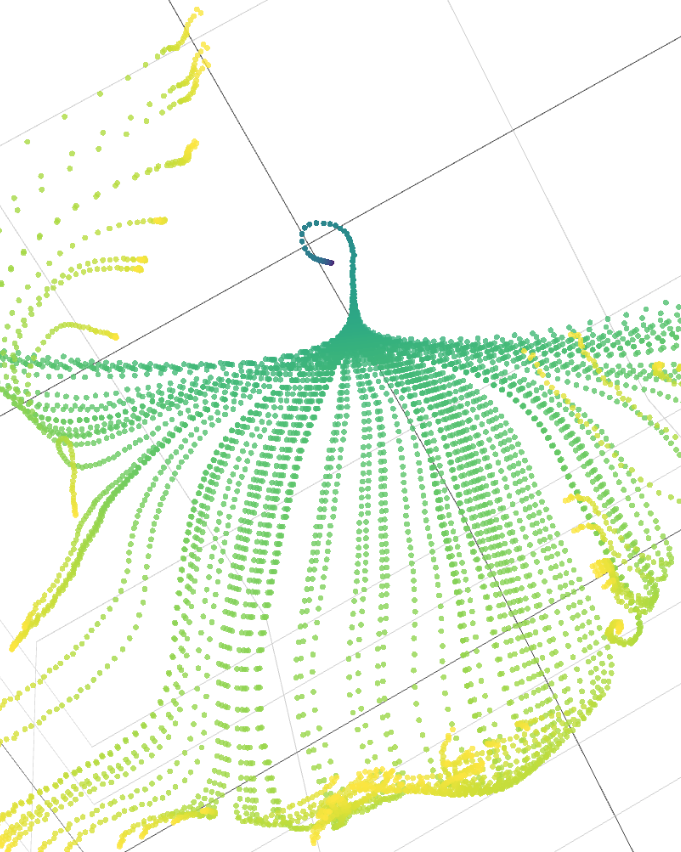}
  \hfill
  \includegraphics[width=0.49\linewidth]{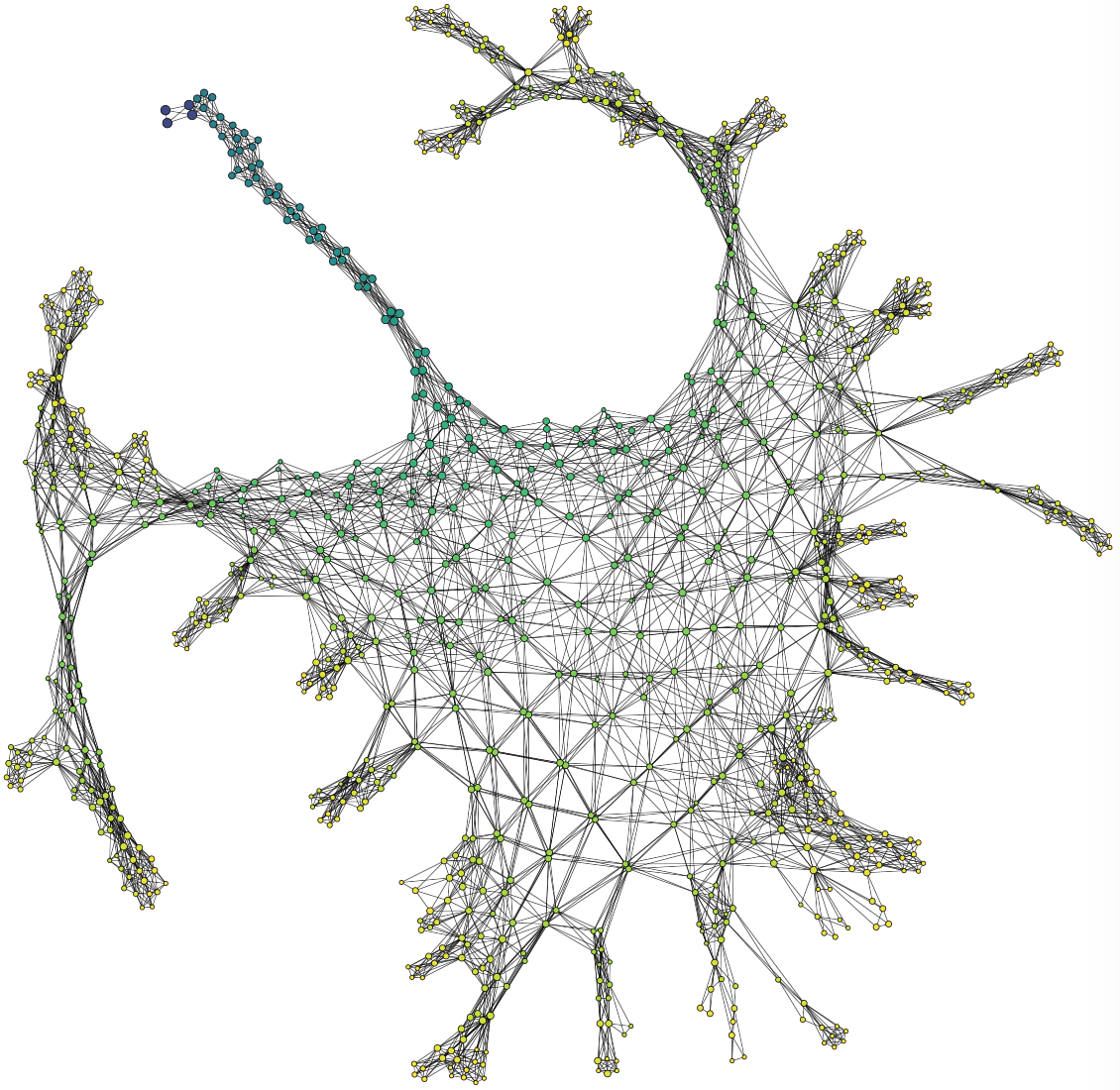}
  \caption{\emph{Left}: Detail of the PCA projection of the evolution of the weights for the first hidden layer, as in the top left of Figure~\ref{Tiny}. Several phases are visible: uniform evolution (blue), parallel evolution along a surface (green), chaotic evolution (yellow).
  \emph{Right}: learning graph representing the surface as a densely connected grid (filter function given by the first 3 PCA directions). 
}
  \label{TinySurface}
\end{figure}

\begin{figure}
  \centering
  \begin{overpic}[width=0.9\textwidth]{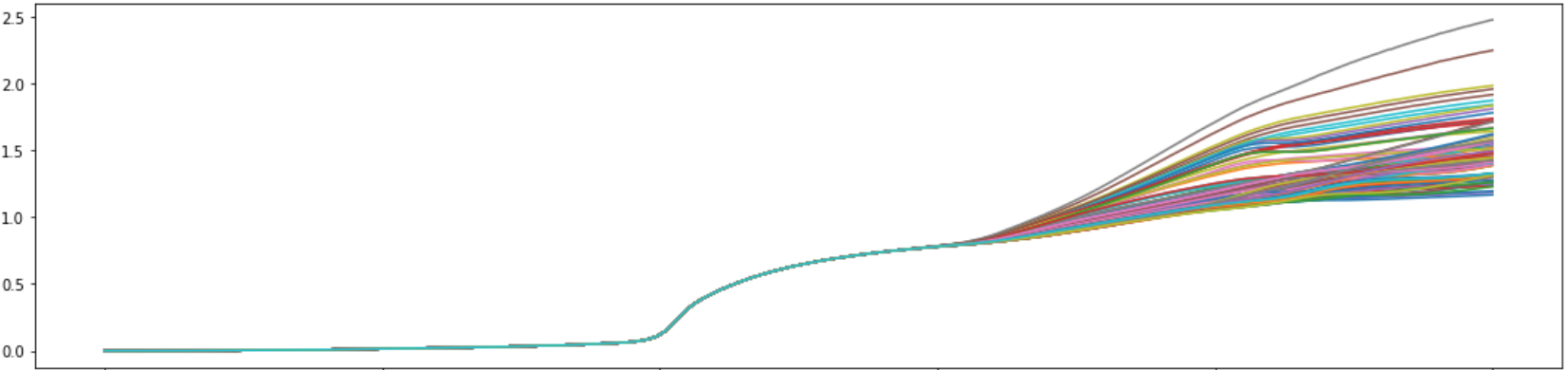}
    \small\put (15,21) {Hidden layer 1}
  \end{overpic}
  \hfill
  \begin{overpic}[width=0.9\textwidth]{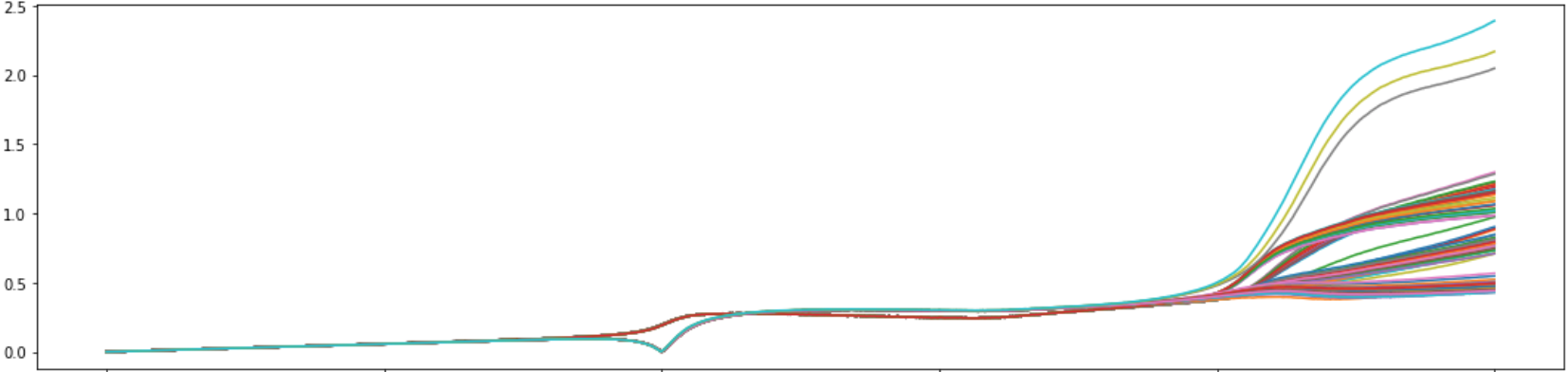}
    \small\put (15,21) {Hidden layer 2}
  \end{overpic}
  \hfill
  \begin{overpic}[width=0.9\textwidth]{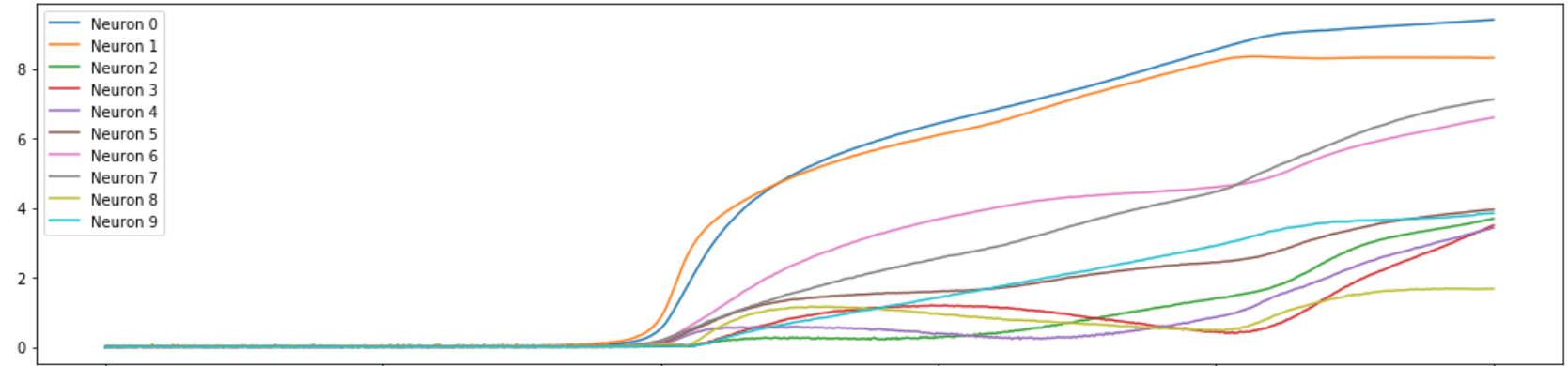}
    \small\put (15,21) {Output layer}
  \end{overpic}
  \hfill
  \begin{overpic}[width=0.9\textwidth]{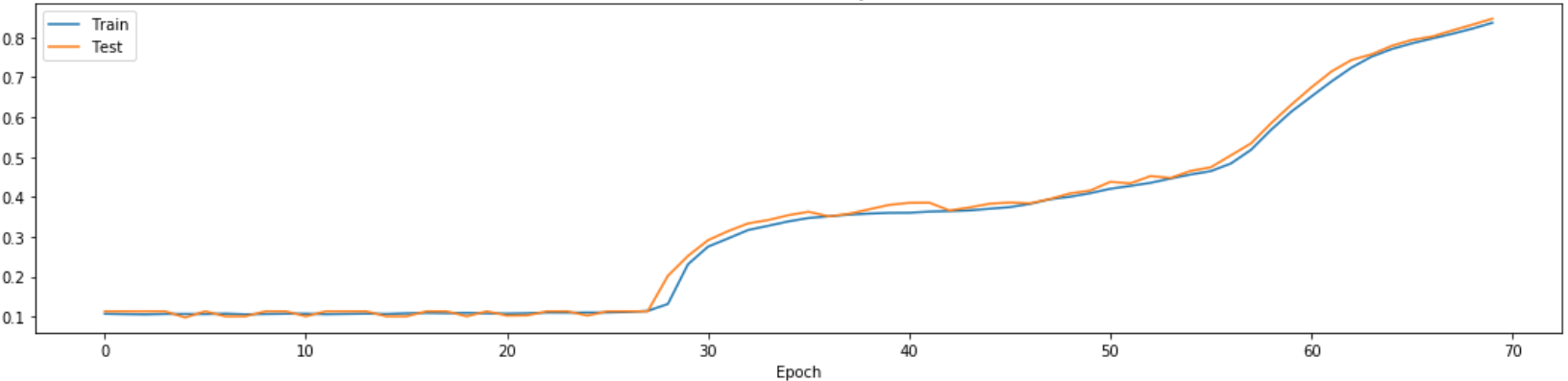}
    \small\put (15,21.8) {Accuracy}
  \end{overpic}
  \caption{\emph{Top}: Norms of the weights per neuron for the two hidden layers and the output layer. The uniform increase is triggered by branching in the output layer. The branching is followed by a short period of reshuffling, before a converging period. 
  \emph{Bottom}: The accuracy increases in jumps: 10\% (chance), 40\%, 80\% (with more training steps, this would be followed by a slow increase towards 95\%).
  }
  \label{NormsAcc}
\end{figure}

The weights of the output layer follow a similar evolution to the case with zero initialization. A phase of slow progression away from the origin is followed by a branching phase and the development of a tree with 10 distinctive branches, associated with the 10 digit classes. The behavior of the weights for the first hidden layer is more surprising (Figure~\ref{TinySurface} left): after following the same trajectory for a while, all the weights branch off simultaneously and appear to evolve in parallel along a \emph{smooth surface}! This coordination appears to break off eventually, and the weights veer away from the surface in arbitrary directions. It is however possible that this apparently chaotic phase is an artifact of the PCA projections, and that the weights in fact continue to evolve along a smooth surface in higher dimensions (projecting a curtain to the floor would give a noisy curve\ldots). This interpretation is supported by the smooth evolution of the weight norms towards the end of training (Figure~\ref{NormsAcc}).

The corresponding learning graph is again a tree, in which the surface has been decomposed into branches. This is due to the fact that with the $L^2$ norm as filter function the edges of the graph are mostly created in the radial direction from the origin, rather than in the lateral direction of the surface. In order to better exhibit the surface we can instead take the filter function to be the first 3 PCA directions. The learning graph then contains a densely connected grid that corresponds to the surface (Figure~\ref{TinySurface} right). With an implementation of the Mapper algorithm taking into account 2-simplices, this would look like a triangulated surface.

Further experimentations indicated that this learning surface is remarkably robust under variations of the training parameters.
While the shape of the surface could have been expected to be determined by the exact random values of the initial weights, it appears in fact to be independent of the random seed. Moreover, the surface varies continuously under small changes in the number of neurons, learning rate, batch size, or mean $\mu$ and standard deviation $\sigma$ of the normal initialization. Comparing the cases of neural networks with one and two hidden layers (Figures~\ref{ZeroInit} and \ref{Tiny}) we also observe a striking similarity between the weight trajectories for the last two layers.

\begin{figure}[b!]
  \centering
  \includegraphics[width=0.83\linewidth]{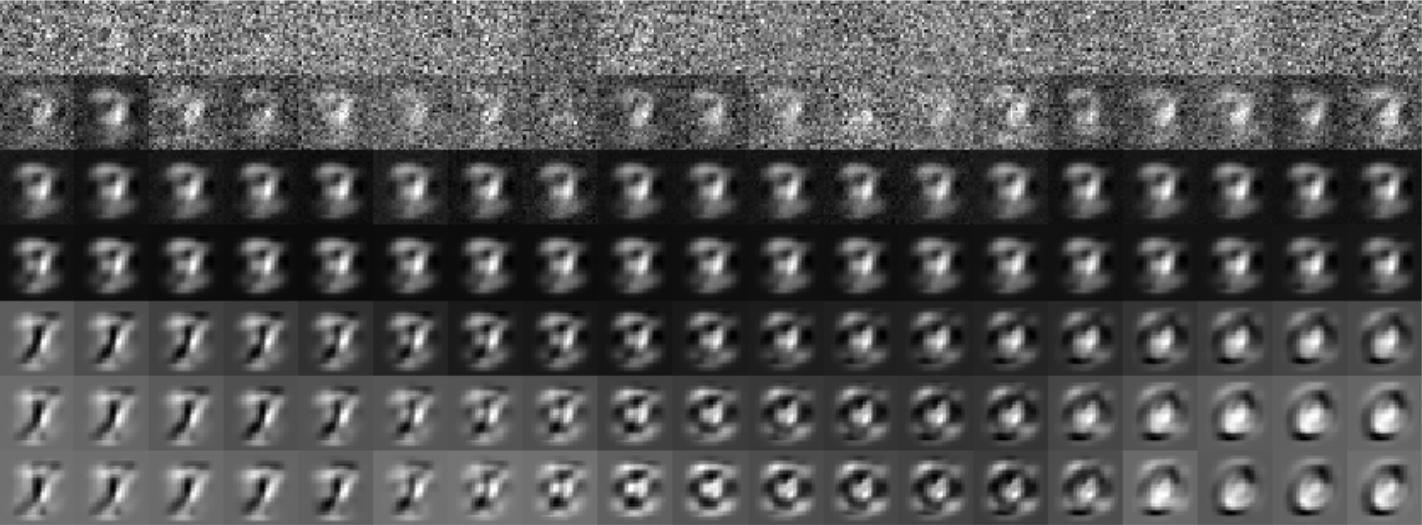}
  \caption{Natural coordinates on the learning surface from Figure~\ref{TinySurface}, provided by differences of the weights of nearby neurons. The downward direction is the training time; the horizontal direction is the lateral spread of neurons across the surface. The last row shows a continuous transformation between the shapes of the digits 1, 3, and 0.}
  \label{Coordinates}
\end{figure}

One idea to try to understand the meaning of the learning surface is to study deformations of the weights as we move on it. Natural coordinates on the surface are provided by the training time and the lateral spread of the diversifying neurons. Sampling neurons along these two dimensions, we can reshape the corresponding weight vectors into $28\times28$ images and study the factors of variation. The images themselves appear rather homogenous, but taking differences between nearby images (for each training step) reveals interesting patterns (Figure~\ref{Coordinates}). As the training progresses, shapes of digits emerge from white noise. For a fixed training step, moving across the surface corresponds to morphing digits into one another, for example a 1 into a 3 into a 0  as in the last row of Figure~\ref{Coordinates}. 
This should be compared with the 2d latent manifolds learned with variational auto-encoders in~\cite{2013arXiv1312.6114K}. It is therefore tempting to conjecture that the learning surface is shaped to capture dominant factors of variation. 

Finally, we turn to the evolution of the weights of the second hidden layer (Figure~\ref{Tiny} middle). The weights first evolve slowly all together along a line, but then they split into two groups, one of which shoots off in the opposite direction, back to the origin and beyond. The two groups of weights undergo a second phase of slow evolution on opposite ends of the line. Branching takes place relatively late and produces surfaces shaped like palm leaves. This case is thus some hybrid between a tree and a learning surface. 

The evolutions of the different layers can be easily compared by looking at plots of the norms of the weights (Figure~\ref{NormsAcc}). During the phase of slow evolution (epochs 0-27), the most noticeable progress is a steady increase in the norms of the weights of the second hidden layer. Branching between the 10 neurons of the output layer then takes place around epoch 28, accompanied by a sharp increase in the weights of the first hidden layer, and the characteristic binary splitting of the weights in the second hidden layer (curiously, the blue curve looks like minus the derivative of the red one). Correspondingly, the accuracy of the model, which was so far attributing all test images to a single class, jumps up to around 40\%. Further inflections in the accuracy occur when the neurons of the first hidden layer spread out into the learning surface (epoch 45), and when the weights of the second hidden layer branch off into palm leaves (epoch 56). The training was interrupted earlier in order to focus on the smooth part of the learning surface, but the accuracy would have otherwise steadily increased towards 96\% as the weight norms converged. 

\section{Discussion}

We have presented a method to visualize the evolution of weights during the training of a fully connected feedforward neural network. It would be desirable to obtain a better theoretical understanding of the various phenomena that we have observed. The remarkable regularity of weight trajectories begs the question of how they are determined and what triggers branching, perhaps from the perspective of bifurcation theory. Similarly, the properties of learning surfaces deserve further study, for example by defining some Riemannian metric on them or by representing them as algebraic curves. Furthermore, the surprising appearance of two-dimensional structures in parameter spaces with hundreds of dimensions could provide some clues to explain the uncanny generalization aptitudes of neural networks. Constraining the weights to evolve on a surface may indeed act as an implicit regularization~\cite{neyshabur2014search}.

Extensions of our work include studying how learning graphs are impacted by initialization schemes, optimization methods, regularization, biases, activation functions, and depth. It would also be interesting to explore learning graphs for convolutional and recurrent neural networks. 

While our main hope was to gain insights into learning in neural networks, our approach can also have practical applications, for example in guiding the choice of architecture and hyper-parameters. We can imagine that quantitative properties of learning graphs, such as the degrees or centralities used in network science, could provide reliable measures to improve the quality of the learning. 

\subsubsection*{Acknowledgments}
We thank Alex Cole, Martin Jaggi, Giovanni Petri, and Dan Roberts for insightful discussions, and Nitya Afambo for early collaboration.

\medskip

\end{document}